\documentclass[10pt,twocolumn,letterpaper]{article}

\usepackage[pagenumbers]{cvpr} 

\usepackage[dvipsnames]{xcolor}

\definecolor{cvprblue}{rgb}{0.21,0.49,0.74}
\usepackage[pagebackref,breaklinks,colorlinks,citecolor=cvprblue]{hyperref}
\usepackage[utf8]{inputenc} 
\usepackage[T1]{fontenc}    
\usepackage{hyperref}       
\usepackage{url}            
\usepackage{booktabs}       
\usepackage{amsfonts}       
\usepackage{nicefrac}       
\usepackage{microtype}      
\usepackage{xcolor}         
\usepackage{multirow}
\usepackage{amsmath}
\usepackage{graphicx}
\usepackage{subcaption}
\newtheorem{theorem}{Theorem}
\usepackage{enumitem}
\usepackage[linesnumbered, boxed, ruled]{algorithm2e}
\SetCommentSty{mycommfont}
\usepackage{wrapfig,lipsum,booktabs}

\def\paperID{13272} 
\def\confName{CVPR}
\def\confYear{2024}

\title{SAVE: Spectral-Shift-Aware Adaptation of Image Diffusion Models for Text-driven Video Editing}

\author{First Author\\
Institution1\\
Institution1 address\\
{\tt\small firstauthor@i1.org}
\and
Second Author\\
Institution2\\
First line of institution2 address\\
{\tt\small secondauthor@i2.org}
}

\begin{document}

\author{%
Nazmul Karim$^*$  \ Umar Khalid$^*$ \ Mohsen Joneidi \ Chen Chen \ Nazanin Rahnavard  
\\
\\
University of Central Florida 
}

\maketitle

{ \renewcommand{\thefootnote}%
    {\fnsymbol{footnote}}
  \footnotetext[1]{Equal Contribution}
}

\begin{figure*}[t]
\centering
    \includegraphics[width=0.8\linewidth, trim={0cm 0cm 0cm 0cm}]{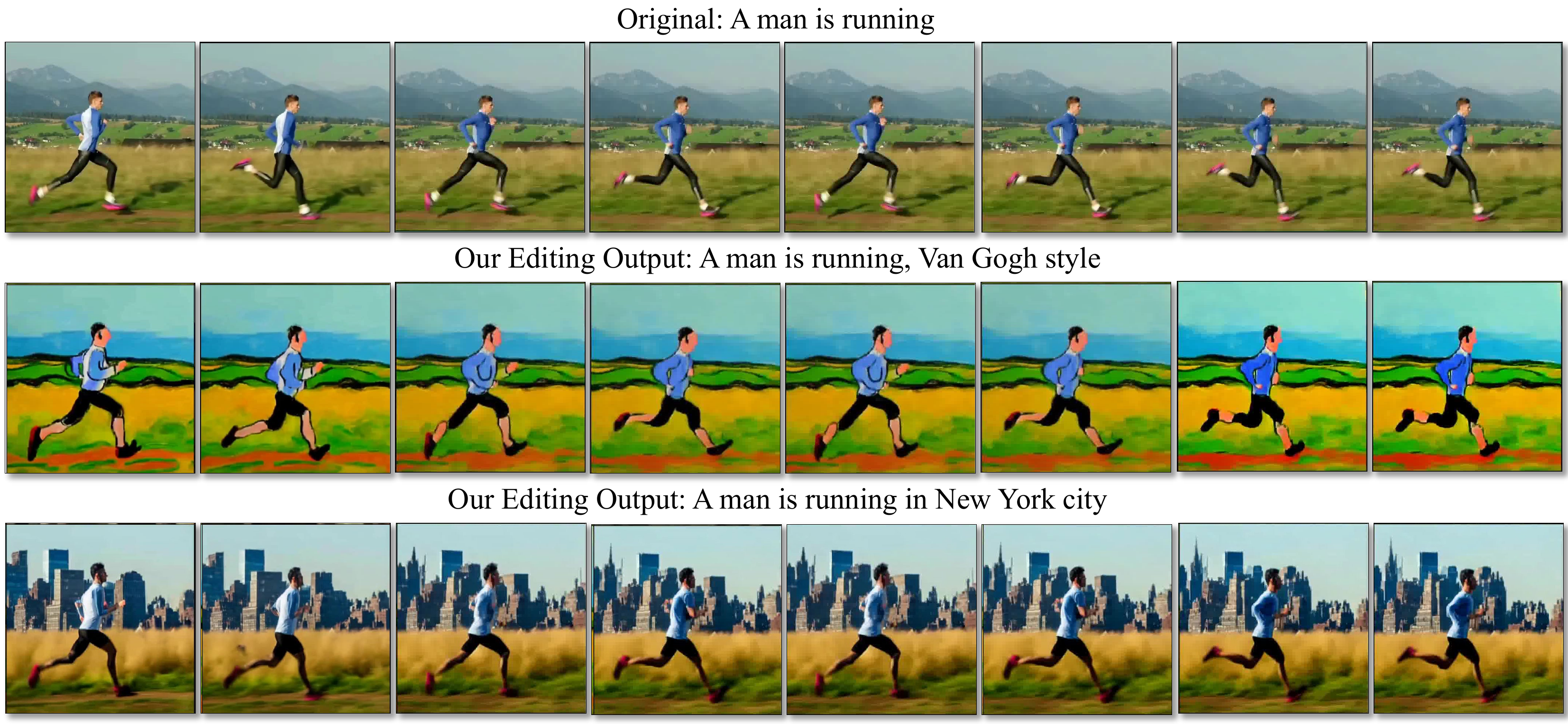}
    \vspace{-1.5mm}
    \caption{\footnotesize Our proposed method \textit{SAVE} enables text-based video editing (e.g. shape, style, etc.) by instilling both spatial and temporal awareness into image diffusion models. }
\label{fig:teaser}
\vspace{-3mm}
\end{figure*}

\begin{abstract}
Text-to-Image (T2I) diffusion models have achieved remarkable success in synthesizing high-quality images conditioned on text prompts. Recent methods have tried to replicate the success by either training text-to-video (T2V) models on a very large number of text-video pairs or adapting T2I models on text-video pairs independently. Although the latter is computationally less expensive, it still takes a significant amount of time for per-video adaption. To address this issue, we propose $\textit{SAVE}$, a novel spectral-shift-aware adaptation framework, in which we fine-tune the spectral shift of the parameter space instead of the parameters themselves. Specifically, we take the spectral decomposition of the pre-trained T2I weights and only update the singular values while freezing the corresponding singular vectors. In addition, we introduce a spectral shift regularizer aimed at placing tighter constraints on larger singular values compared to smaller ones. This form of regularization enables the model to grasp finer details within the video that align with the provided textual descriptions. We also offer theoretical justification for our proposed regularization technique. Since we are only dealing with spectral shifts, the proposed method reduces the adaptation time significantly ($\sim 10\times$) and has fewer resource constraints for training. Such attributes posit $\textit{SAVE}$ to be more suitable for real-world applications, e.g. editing undesirable contents during video streaming. We validate the effectiveness of $\textit{SAVE}$  with an extensive experimental evaluation under different settings, e.g. style transfer, object replacement, etc. 


\end{abstract}

\section{Introduction}
\vspace{-1mm}
Diffusion models~\cite{ddpm} have shown tremendous success in the text-guided synthesis of diverse and high-quality media contents such as images~\cite{imagen, dalle2} and videos~\cite{ho2022imagen, singer2023makeavideo, zhou2022magicvideo, lvdm}. Due to the strong data modeling capabilities of these models, diversified generation~\cite{rombach2022high} of a wide range of objects, shapes, and styles has become possible with remarkable realism. In recent times, several diffusion-based editing methods~\cite{avrahami2022blended,avrahami2022blendedlatent,couairon2022diffedit,pix2pix-zero,p2p,pnp} also made their way into generative AI research. For example, customizable or personalized image diffusion models~\cite{ruiz2022dreambooth,kumari2022multi} leverage parameters fine-tuning for adapting the model to user-specific editing requirements, e.g. shape, style, etc. Although image editing has earned its popularity, we focus on designing an~\emph{efficient text-guided video-editing framework}.

In general, there are two primary ways for video generation: 1) training a T2V diffusion model on a large-scale multimodal (text-video pairs) dataset~\cite{ho2022cascaded,ho2022classifier,ho2022imagen,singer2023makeavideo} and 2) modifying the existing T2I diffusion models to fit the video generation process~\cite{wu2022tune,text2live}. The latter garnered more interest due to the inherent challenges associated with acquiring a large-scale text-video dataset as compared to a text-image dataset. Furthermore, it is computationally expensive to train a video model as it requires a larger parameter space (having to accommodate temporal dimension) compared to a T2I model. Consequently, adapting image diffusion models presents a more feasible option due to their widespread availability~\cite{mou2023t2i,zhang2023adding,rombach2022high}. Although generative priors from a T2I model supplement the spatial component of the video generation process, they lack temporal awareness making it harder to model the motion and 3D shape understanding. Tune-A-Video~\cite{wu2022tune} and Text2LIVE~\cite{text2live} tried to address this by adding temporal layers to the T2I model for instilling temporal awareness. Despite the promising results, the issue of computational overhead is prominent as we still have to fine-tune a large number of parameters. There are two critical factors involved in diffusion-based video generation or editing: accurately reconstructing the original video based on the source prompt and ensuring easy modification in line with the target prompt. Note that, we are employing an T2I model and adapting it to be a T2V model. The final T2V model is not trained on millions of videos, but rather only one. As a result, it is not as generalizable as the T2I model. Therefore, to have better editability T2V needs to preserve much of the generalization capability of T2I.
 

To address these issues, we propose to tune the spectral shift of the parameter space such that the underlying motion concept, as well as content information in the input video, is learned without compromising the generalization ability of the original T2I model. For spectral shift tuning, we first take the singular value decomposition (SVD) of the pre-trained weights from each layer and freeze the singular vectors while updating only the singular values iteratively. We also notice that unconstrained optimization of spectral shifts can lead to drastic updates of larger singular values which can be catastrophic considering the goal here is to learn fine details of the video that matches the description of a text prompt. As a remedy, we propose a novel \emph{spectral shift regularizer} that allows minimal changes to larger singular values. In addition, we conduct a comprehensive study of different spatiotemporal attentions for video editing and propose single \emph{frame attention} for better computational efficiency. To this end, we propose a novel \textbf{\textit{S}}pectral shift \textbf{\textit{A}}ware  \textbf{\textit{V}}ideo \textbf{\textit{E}}diting (\textbf{\textit{SAVE}}) technique that only fine-tunes the spectral shift of the parameter space for efficient adaptation. By tuning only the singular values, we reduce the trainable parameters significantly and speed up the adaptation proportionally almost \textbf{10$\times$}. Our contributions can be summarized as follows:

\begin{itemize}
    \item We propose a novel text-guided video editing framework that adapts an image diffusion model by only fine-tuning the spectral shift of its parameter space. This allows us to have a significantly reduced number of tunable parameters with better computational efficiency. To the best of our knowledge, we are the first to address the video generation problem from the spectral shift perspective.

    \item A spectral shift regularizer is introduced with the aim of placing tighter constraints on larger singular values compared to smaller ones. Such regularization facilitates the model to learn finer details in the video that match the given text descriptions. We also provide theoretical justification for our proposed regularizer.     

    \item Based on our comprehensive study, we propose to incorporate a frame attention mechanism that enforces spatial and temporal consistency across the frames and also offers better efficiency. We extensively evaluate the effectiveness of our method in different benchmarks with both qualitative and quantitative results; a snapshot of which is shown in Fig.~\ref{fig:teaser}.  
\end{itemize}

\begin{figure*}[t]
    \centering
    \includegraphics[width=0.9\textwidth]{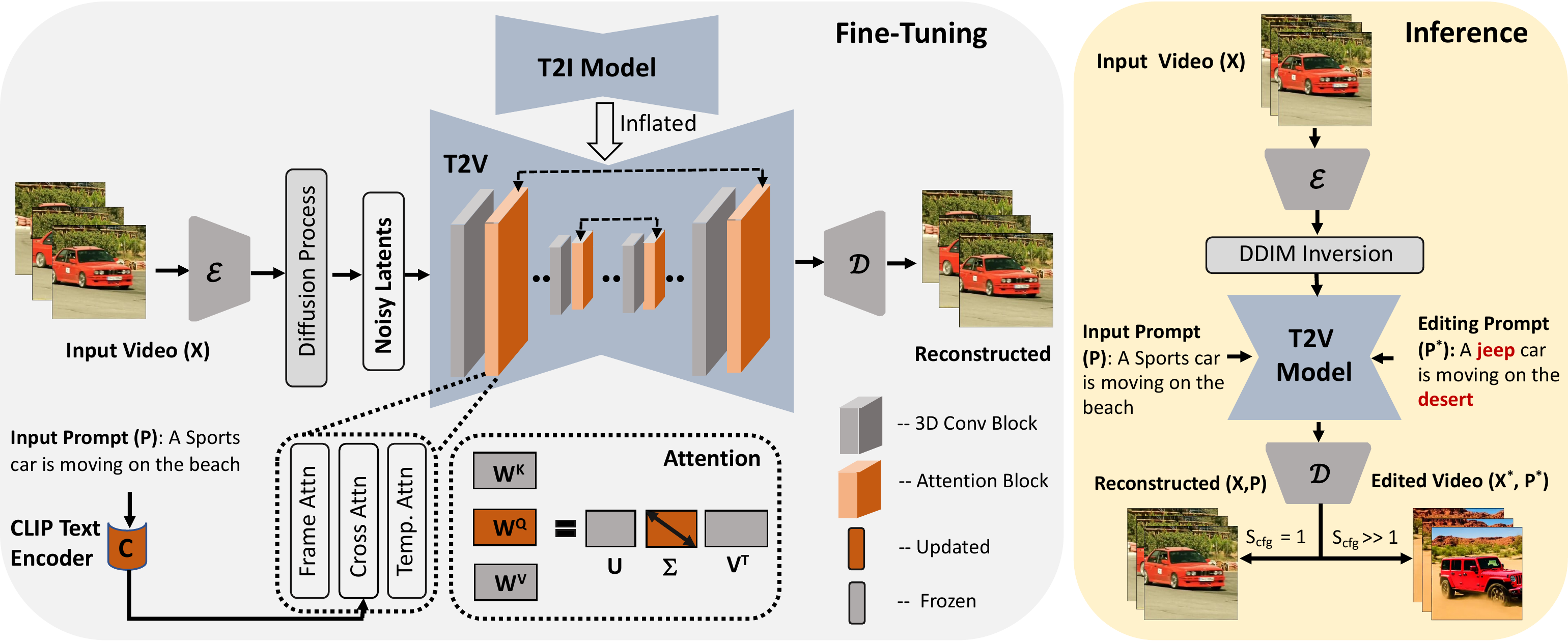}
    \caption{\footnotesize \textbf{An illustration of our text-to-video editing framework}. We initialize our T2V model with a pre-trained inflated T2I model, where we repeat the weights in the temporal dimension. We feed the denoising T2V model with an input video ($X$) and a corresponding prompt ($\mathcal{P}$) before converting the clean latents into noisy latents using forward diffusion process. Along with cross-attention, we have a frame attention layer for learning motion information as well as content consistency over frames. We only update the weights corresponding to query ($W^{Q}$). However, instead of updating all parameters, we merely fine-tune the singular values ($\Sigma = diag(\sigma)$). This reduces the number of trainable parameters significantly. In addition, we propose to regularize $\Sigma$ to match the fine details of $X$ to the text description in $\mathcal{P}$ and obtain superior performance in significantly less time. At inference, we pass the editing prompt ($\mathcal{P^\ast}$) that warrants the required changes in the output video. Depending on the classifier-free guidance scale $s_{cfg}$, the editing results may vary. With $s_{cfg}=1$, we should be able to recover the original video.  }
    \vspace{-3mm}
    \label{fig:summary}
\end{figure*}

\vspace{-1mm}
\section{Related Work}
\vspace{-1mm}


\noindent \textbf{Text-to-Image Diffusion Models.} The field of Text-to-Image (T2I) generation has been extensively investigated, with many models based on transformers being proposed in recent years~\cite{ramesh2021zero, yu2022scaling, yu2021vector, ding2022cogview2, gafni2022make}. In an effort to enhance the quality of generated images, several T2I generative models have incorporated diffusion models~\cite{ddpm}. For example, GLIDE incorporated classifier-free guidance within the diffusion framework to enhance image quality~\cite{nichol2021glide}, while DALLE-2 improved text-image alignments through the utilization of CLIP feature space~\cite{dalle2}. Imagen employed cascaded diffusion models for generating high-definition videos~\cite{ho2022imagen}, and subsequent works such as VQ-diffusion and Latent Diffusion Models (LDMs) operated in the latent space to enhance training efficacy~\cite{gu2022vector, rombach2022high}. 

\medskip

\noindent \textbf{Text-to-Video Generative Models.} While remarkable progress has been made in text-to-image (T2I) generation, the field of generating videos from text prompts still lags behind. This is primarily due to the limited availability of large-scale text-video datasets and the inherent challenges in modeling temporal consistency and coherence. Early works in this domain, such as VideoGAN~\cite{li2018video}, ImagineGAN~\cite{gupta2018imagine}, and CrossNet~\cite{liu2019cross}, primarily focused on  generating simple videos, such as moving digits or specific human activities. More recently, GODIVA~\cite{van2017neural} introduced a model that utilizes a 2D Vector Quantized Variational Autoencoder (VQ-VAE) with sparse attention for text-to-video (T2V) generation, enabling more realistic scene synthesis. 
Video Diffusion Models (VDM)~\cite{ho2022video} builds upon the advancements of T2I models by employing a space-time factorized U-Net architecture and training with both image and video data, thereby achieving improved performance in video generation tasks.~\cite{ho2022imagen} further enhanced VDM by employing cascaded diffusion models to generate high-definition videos. Make-A-Video~\cite{singer2023makeavideo} and MagicVideo~\cite{zhou2022magicvideo} pursued similar goals of transferring progress from T2I generation to T2V generation. Recently, a few LDM stable diffusion-based methods~\cite{an2023latentshift, blattmann2023align,he2023latent} are proposed for efficient video generation. In our approach, we extend the capabilities of LDMs by expanding the 2D model into the spatiotemporal domain within the latent space that efficiently fine-tunes pre-trained T2I diffusion models on a single text-video pair.


\medskip

\noindent \textbf{Text-Driven Video Editing.} With the success of diffusion-based image editing works~\cite{shi2023instantbooth,chen2023disenbooth,liu2022compositional,avrahami2022blended,avrahami2022blendedlatent,kawar2022imagic,zhang2022sine,orgad2023editing,wallace2022edict,pnp,pix2pix-zero,bansal2023universal,wallace2023end, ruiz2022dreambooth}, a few diffusion-based video-editing frameworks have been proposed.  Dreamix~\cite{molad2023dreamix}, Gen-1~\cite{esser2023structure}, and Tune-A-Video~\cite{wu2022tune} either employ VDM or leverage the pre-trained T2I models for video editing. Although these approaches have shown impressive results, it is important to note that VDMs are computationally challenging and require large-scale captioned images and videos for training. One-shot T2I-based video editing methods such as Tune-A-Video~\cite{wu2022tune} employ the model inflation technique and fine-tunes the temporal attention weights. However, their editing capabilities are limited by the pre-trained T2I models while our work opens up new avenues for the efficient and effective fine-tuning text-to-image diffusion models for personalization and customization for video editing tasks. 
\section{Method}
\vspace{-1mm}

Let $\mathcal{X}=\{x_i|i\in[1,F]\}$ be a video comprising $F$ frames and $\mathcal{P}$ be the input prompt that describes the content in $\mathcal{X
}$. Our objective is to generate a novel video $\mathcal{X^\ast}$ with editing commands coming from the prompt $\mathcal{P^\ast}$. Although a pre-trained Text-to-Video (T2V) diffusion model can be employed to edit $X$, training such a model can be computationally expensive~\cite{ho2022cascaded,ho2022classifier,ho2022imagen,singer2023makeavideo}. 
In this paper, we propose a novel video editing technique, $\textit{SAVE}$, that achieves the same objective by leveraging a publicly available pre-trained Text-to-Image (T2I) model and a single text-video pair. In the following, we provide a brief background on diffusion models in Section~\ref{sec:preliminary}, followed by the details of our proposed method in Section~\ref{sec:sepctral_shift}. An overview of our framework is illustrated in Fig.~\ref{fig:summary}.
\vspace{-1mm}
\subsection{Preliminaries}\label{sec:preliminary}
\vspace{-1mm}

\paragraph{Diffusion Models.} Stable diffusion (SD)~\cite{rombach2022high} model operates on the latent space of an autoencoder $\mathcal{D}(\mathcal{E}(\cdot))$, namely VQ-GAN \cite{esser2021taming} or VQ-VAE \cite{van2017neural}. Here, $\mathcal{E}$ is the encoder that compresses an RGB image $x$ to a low-resolution latent $z = \mathcal{E}(x)$, which can be recovered using the decoder $x \sim \mathcal{D}(z)$. The diffusion forward process iteratively adds Gaussian noise to the signal $z$: 
\begin{equation}
    q(z_t|z_{t-1}) = \mathcal{N}(z_t; \sqrt{1-\beta_t}z_{t-1}, \beta_t I), \;
    t = 1,2\ldots,T,
\end{equation}
where $q(z_t|z_{t-1})$ is the conditional density of $z_t$ given $z_{t-1}$, $\{\beta_t\}_{t=1}^{T}$ are hyperparameters.
$T$ is chosen large enough such that $z_T \sim \mathcal{N}(0,I)$. After getting the noisy latents $\{z_t; t = 1,2,\ldots,T\}$, a U-Net~\cite{ronneberger2015u} composed of convolutional as well as self and cross attentional blocks with parameters $\theta$  is trained for the backward process, a.k.a denoising, using the objective function:
\vspace{-0.5em}
\begin{equation}
\min _\theta E_{z_0, \varepsilon \sim N(0, I), t \sim \text { Uniform }(1, T)}\left\|\varepsilon-\varepsilon_\theta\left(z_t, t, p \right)\right\|_2^2,
\end{equation}
where $p$ is the embedding of prompt $p = \mathcal{C}(\mathcal{P};\phi)$ and $\varepsilon_\theta$ is the model predicted noise at time $t$. During inference, we apply deterministic DDIM sampling~\cite{ddim} to convert a random noise $z_T$ to a clean latent $z_0$. We move the discussion regarding DDIM Sampling and Inversion to the \emph{\textcolor{blue}{Supplementary}}.

\vspace{-1.5mm}
\paragraph{Text-driven Video Generation.}  To perform editing operations, one needs to find the variables in the latent space that corresponds to the frame contents. After that, we can edit contents by finding editing directions in the latent space. The editing direction is usually provided by $\mathcal{P}^*$ while setting $s_{cfg}>>1$.  While using a large $s_{cfg}$ gives more freedom in editing, this freedom can also lead to frame inconsistency. Furthermore, the issue of error accumulation can also cause such inconsistency given that we consider 50 DDIM inversion steps. These issues are less prominent once we fine-tune the T2V model with a text-video pair ($X$,$\mathcal{P}$) that aligns the text embedding with the video content. To obtain better alignment, we also fine-tune the text encoder $\mathcal{C}$ for improved text-video alignment.         

\begin{figure}
\centering
    \vspace{-3mm}
    \includegraphics[width=0.9\linewidth, trim={0cm 1cm 9cm 1cm}]{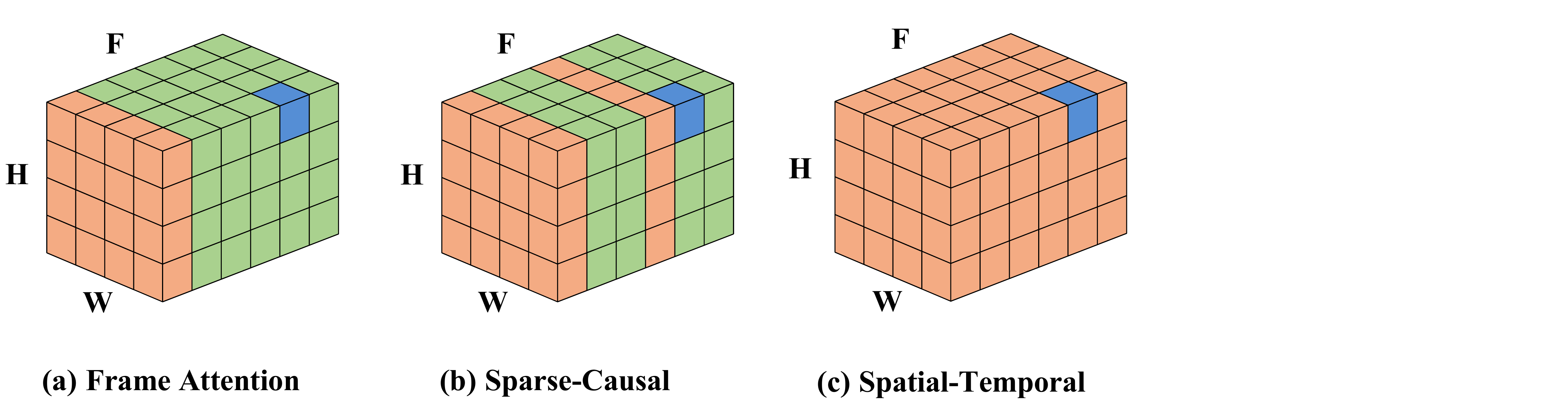}
    \caption{\footnotesize{Showcasing various \textbf{cross-frame attention mechanisms}. In the visual representation, the query and keys are denoted by the color orange and blue, respectively. The variables H, W, and F correspond to the height, width, and number of frames in the input video, respectively.}  }
    \vspace{-1mm}

\label{fig:attn_comp}
\end{figure}
\begin{table}
\captionsetup{type=table} 
    \centering

    \scalebox{0.75}{
    \begin{tabular}{c|c}
    \toprule
    Attention Mechanism & Attention FLOPs/Unet Block \\
    \midrule
    Spatio-temporal & $Attention_{c}\times {F\times H} $
    \\
    Sparse-causal+Temopral& $Attention_{c}\times {[F + 2H]} $\\
   
    Frame+Temopral & $Attention_{c}\times {[F + H]} $
    \\
    \bottomrule
    \end{tabular}
    }    
    \vspace{-1.5mm}
    \caption{\footnotesize \textbf{FLOPs comparison} of various attention variants for Unet blocks. Here,  $Attention_{c}=4\times B^{'}\times F\times H \times W$, and $B^{'}=$ Batch Size $\times \#$ of attention heads. In our settings, we use frame attention along with the temporal-attention.} 
    \label{tab:atten_flops}
    \vspace{-3.5mm}

\end{table}

\subsection{Spectral-Shift-Aware Video Editing (SAVE)}\label{sec:sepctral_shift}
For video diffusion, we need to generate $F$ images instead of a single image. However, the pre-trained U-Net model consists of 2D convolution blocks that perform sequential downsampling followed by upsampling passes with skip connections. To process $F$ frames, we follow VDM~\cite{ho2022video} to inflate the 2D convolution to pseudo-3D convolution layers by replacing 2D kernels with 3D kernels. For the attention block, we replace self-attention with spatiotemporal attention layers that take into account information from multiple frames. However, architectural changes are not enough as the model still has to learn the motion information. For that, we need to fine-tune the newly constructed T2V model. In our work, we propose a novel way of fine-tuning by only controlling the spectral shift of the model. 
Following TAV~\cite{wu2022tune}, we only fine-tune the query matrices ($\boldsymbol{W}^{Q}$) of attention layers and freeze all convolution layers. Consider, we have the pre-trained query weight matrices  ($\boldsymbol{W}^{Q}= [\boldsymbol{W}^{Q}_1, \boldsymbol{W}^{Q}_2, ..., \boldsymbol{W}^{Q}_L]$) from a T2I model with $L$ number of attention layers. As shown in Fig.~\ref{fig:summary}, we take spectral decomposition of $\boldsymbol{W}^{Q}_i = \boldsymbol{U_i}\boldsymbol{\Sigma_i} \boldsymbol{V_i}^T \in \mathbb{R}^{M \times N}$, where $\Sigma_i=\text{diag}(\sigma_i)$ and $\sigma_i=[\sigma_i^1, \sigma_i^2, ..., \sigma_i^M]$ are singular values arranged in 
a descending order. Here $M$ is the query embedding dimension of the $i^{th}$ layer.  The spectral shift of the parameter space is defined as the difference between singular values of original $\boldsymbol{W}^{Q}_i$ and the updated $\widehat{\boldsymbol{W}}^{Q}_i$, and can be expressed as $\delta_i = [\delta_i^1, \delta_i^2, ..., \delta_i^{M}]$. Here, $\delta_i^1$ is the difference between individual singular value $\sigma_i^1$. To optimize the spectral shift of our diffusion model, we express the updated singular values as $\boldsymbol{\Sigma}_i^{\delta_i} = \text{diag}(\text{ReLU}(\sigma_i+\delta_i))$ and updated weights $\widehat{\boldsymbol{W_i}} = \boldsymbol{U_i}\boldsymbol{\Sigma}_i^{\delta_i} \boldsymbol{V_i}^T$. The loss function we employ for optimizing the total spectral shift $\delta = [\delta_1, \delta_2, ..., \delta_L]$ is
\begin{align}
    \mathcal{L}(\delta) &= E_{z_0}\left\|\varepsilon-\varepsilon_{\theta_{\delta}}\left(z_t, t, p \right)\right\|_2^2,
\end{align}
where $\theta_{\delta}$ is our diffusion model weights which can be expressed as $\{\delta, \theta_u\}$. Here, $\delta$ is the spectral shift parameters, and $\theta_u$ are the rest of the parameters that were frozen. For example, the parameters related to the K and V are included in $\theta_u$. We observe that uncontrolled optimization of spectral shifts might cause significant updates in larger singular values, which could be problematic as our objective is to capture intricate video details that match a given text prompt. To address this issue, we introduce a novel "spectral shift regularizer" designed to minimize alterations to larger singular values.

\vspace{-1mm}
\subsubsection{Regularized Spectral Shift}\label{sec:regularizer_spectral}
To develop a proper regularizer, we first analyze the data generation process from the perspective of spectral decomposition. To this aim, a new class of linear and simple generators is defined which is referred to as linear spectral generators (LSGs). Next, we will see how this constructive perspective inspires us to develop a regularized spectral shift.


\begin{figure*}[h!]
\centering
    \includegraphics[width=0.75\textwidth, trim={0cm 8.5cm 0cm 0cm},clip]{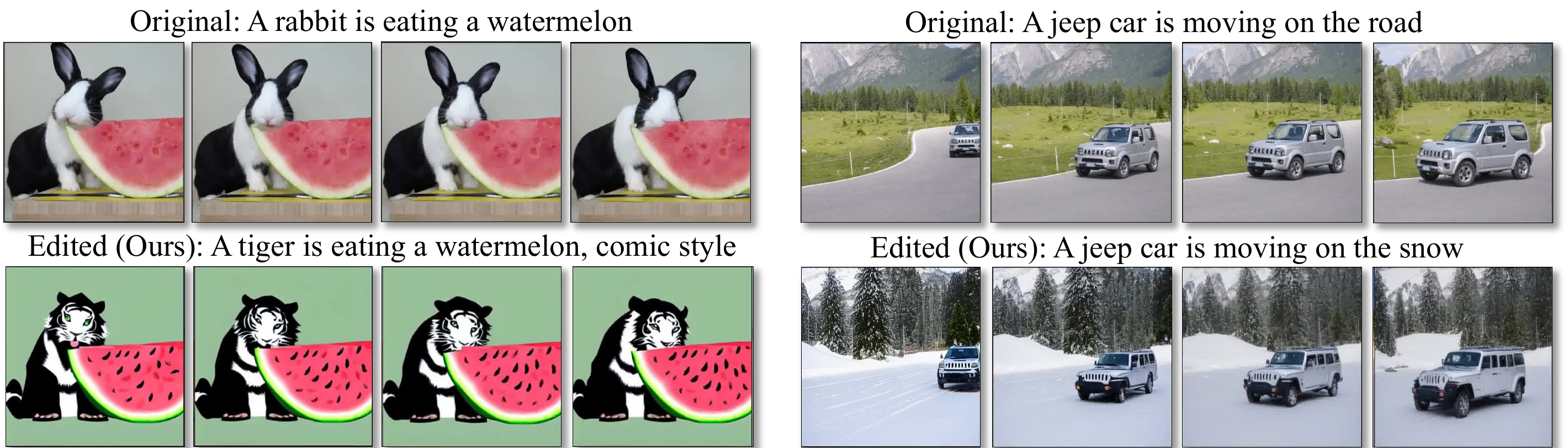}
    \vspace{-2mm}
    \caption{\footnotesize \textbf{Sample editing results} of our proposed method. Zoom in for better visibility.} 
\vspace{-1.5mm}
\label{fig:main_results}
\end{figure*}
\begin{figure*}[t!]
\centering
    \includegraphics[width=0.75\textwidth, trim={0cm 9.3cm 0cm 0cm},clip]{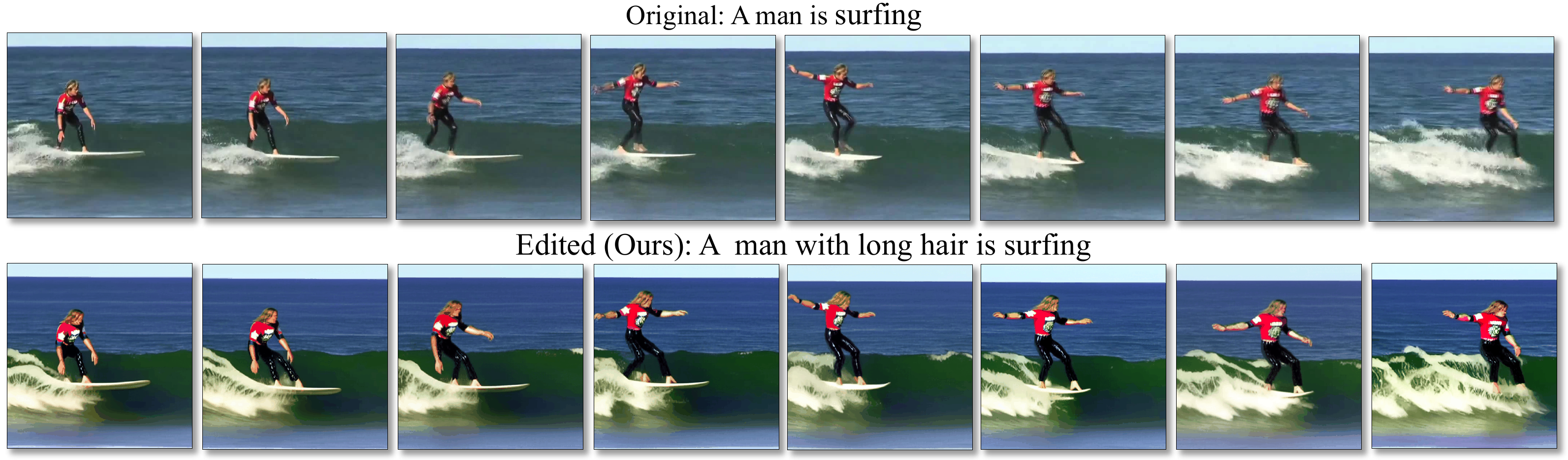}
    \vspace{-2mm}
    \caption{\footnotesize{Transforming object local attributes. Zoom in for better visibility.} }
\label{fig:privacy}
    \vspace{-1.5mm}
\end{figure*}
\begin{figure*}[t!]
\centering
    \includegraphics[width=0.75\textwidth, trim={0cm 0cm 0cm 0cm}]{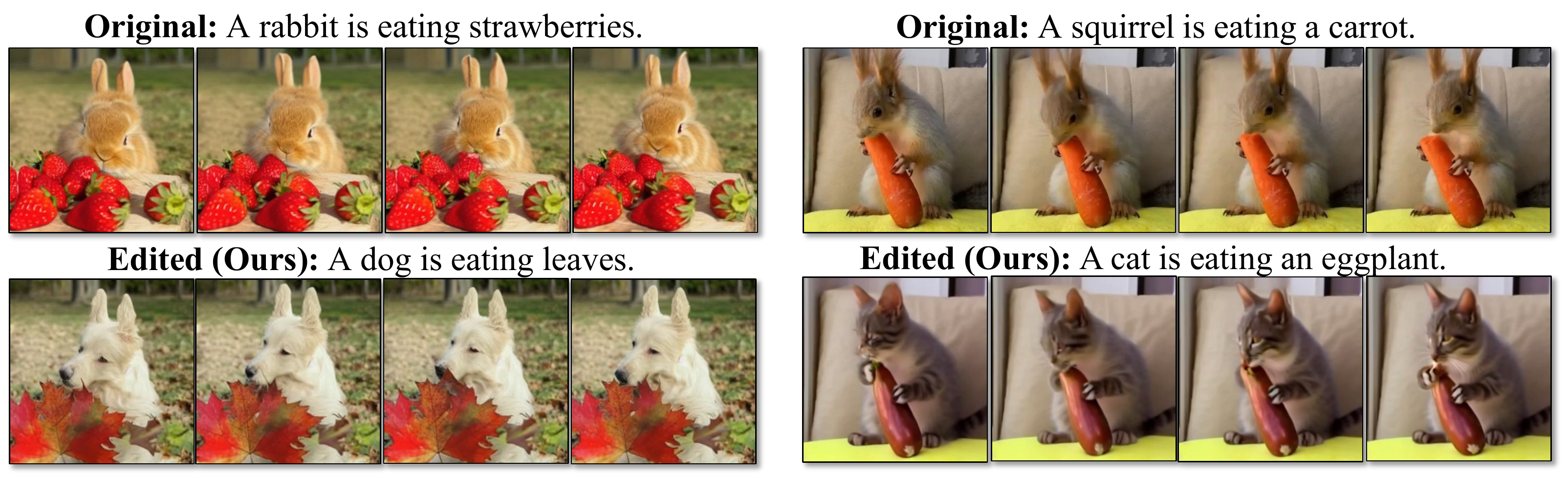}
    \vspace{-2mm}
    \caption{\footnotesize{\textbf{Editing multiple attributes} based on complex prompts. Zoom in for better visibility.} }
\label{fig:multiple_edits}
    \vspace{-2mm}
\end{figure*}

\vspace{-1em}
\paragraph{Linear Spectral Generators.}  Despite the simplicity of linear models, they are useful tools for the definition of more complicated models. For example, convolution is a linear operator and it is well known that a multi-layer of convolutions with non-linear activation functions is capable of performing complicated tasks. Here, we define a basic generator network that can be extended to a deep network, e.g. video diffusion model.

Let $\boldsymbol{D}$ be a matrix whose columns are denoted by $\boldsymbol{d}_i$ $\sim$ $\mathcal{P}_D$, and we are to generate a new sample from the distribution of $\mathcal{P}_D$. 
Let $\boldsymbol{U}\boldsymbol{\Sigma} \boldsymbol{V}^T$ denote SVD of $\boldsymbol{D}$. Let us define $\boldsymbol{C}=\boldsymbol{U}^T \boldsymbol{D}$ as spectral coefficients of the data set, which explains the standard procedure to transform the original domain data $\boldsymbol{D}$ to an intermediate domain or spectral domain. Each row of $\boldsymbol{C}$ shows the contribution of the corresponding spectral component in all samples. The mean and variance of each row point to a normal distribution $\mathcal{N}(m_i,v_i)$.
Here $m_i$ and $v_i$ correspond to mean and variance of the $i^{\text{th}}$ row of $\boldsymbol{C}$, respectively. Drawing a random noise for each spectral component gives us ${\boldsymbol{c}}_g$ which can be transferred back to ${\boldsymbol{d}}_g=\boldsymbol{U}\boldsymbol{c}_g$ as the generated sample.
\vspace{-1mm}
\begin{theorem}\label{thm:singular_value}
    Let $\boldsymbol{D}\in \mathbb{R}^{N\times M}$ and we generate $M$ samples using a linear spectral generator as columns of matrix $\boldsymbol{D}_g$. Then, the absolute difference of the $n^{\text{th}}$ singular value of $\boldsymbol{D}$ and the expected value of the $n^{\text{th}}$ singular value of $\boldsymbol{D}_g$ is less than $\delta_n^{max}$ which is defined as  
\begin{equation}\label{eq:svd}
    \delta_n^{max}  = \frac{2|m_n\sum_{i\ne n} m_i|}{\sigma_n(\boldsymbol{D})+\sigma_n(\boldsymbol{D}_g)}.
\end{equation}
Here, $\sigma_n(\boldsymbol{D})$ and $\sigma_n(\boldsymbol{D}_g)$ are the $n^{\text{th}}$ singular value of $\boldsymbol{D}$ and $\boldsymbol{D}_g$, respectively. 

\textbf{Proof:} Provided in the \textcolor{blue}{Supplementary}.
\end{theorem}

Now, consider we have a one-layer from a pre-trained network ($\boldsymbol{W}^{Q}$) characterized by $\boldsymbol{U}$, $\boldsymbol{\Sigma}$ and $\boldsymbol{V}$ as its spectral components. Then, fine-tuning $\boldsymbol{W}^{Q}$ in the space of spectrally shifted replicas is equivalent to generating $\boldsymbol{W}^{Q}_g (= \widehat{\boldsymbol{W}}^{Q})$ which pursuits singular values of the pre-trained $\boldsymbol{W}^{Q}$ with a uniform maximum distance from the original singular values. However, Theorem 1 suggests we have a regularized neighborhood for each singular value accordingly. This fact encourages us to have a revised loss function for fine-tuning as follows,    
\begin{align}\label{eq:loss_w_reg}
    \mathcal{L}(\delta) &= E_{z_0}\left\|\varepsilon-\varepsilon_{\theta_{\delta}}\left(z_t, t, p \right)\right\|_2^2
    +\lambda\mathcal{L}_{r}(\delta) ,
\end{align}
where $\mathcal{L}_r{(\delta)} = \sum_{i=1}^{L}\delta_i^{T} \Sigma_i \delta_i$ is the regularizer that confines the perturbation ($\delta$) to be more restricted for large singular values and more relaxed for small singular values and $\lambda$ is the regularizer coefficient which is set to be 1e-3.     

\medskip

\begin{figure*}[t!]
\centering
    \includegraphics[width=0.75\linewidth, trim={0.9cm 4.5cm 9.8cm 0cm}]{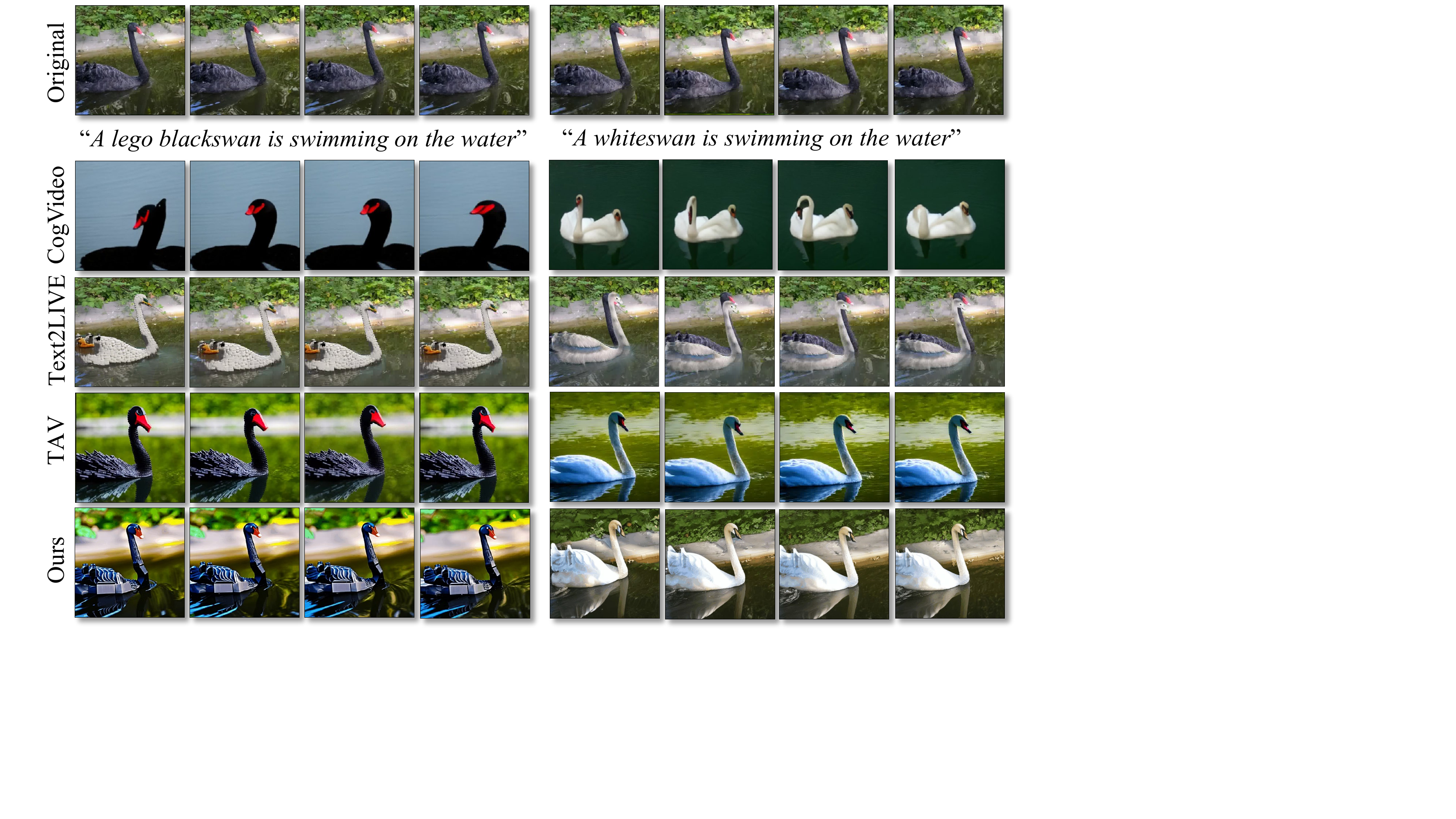}
    \vspace{-2mm}
    \caption{\footnotesize \textbf{Qualitative comparison} between the proposed work and established baselines. }
\label{fig:comparison}
\vspace{-3mm}
\end{figure*}

\noindent \textbf{Implications of Theorem~\ref{thm:singular_value}.} There is a bounding effect on the delta as shown by Theorem~\ref{thm:singular_value} where we argue that the changes in the singular values should be bounded by a certain value that can be derived from the singular values themselves. This is due to the fact that the updated weights $\widehat{\boldsymbol{W}}^{Q}$ (which will generate the edited video) are assumed to be a spectrally perturbed version of the original weights $\boldsymbol{W}^{Q}$. According to Theorem~\ref{thm:singular_value}, these spectral perturbations ($\delta$) can not be random; rather they should be controlled according to the magnitude of singular values $\sigma$. The outcome of this is a compact and regularized parameter space that encourages the model to learn in a more uniform way, i.e. not putting more emphasis on learning a certain part of the prompt or input video frames. Our proposed regularizer restricts main spectral components (i.e. large eigenvalues) to follow the spectrum of original data while it allows small singular values to deviate more from the original spectrum in order to span fine details. This means the first term of Eq.~\ref{eq:loss_w_reg} is responsible for model adaptation to new data while the second term controls the nature of this adaptation. For a given input text-video pair ($X$,$\mathcal{P}$), the model learns the content information of $X$ \emph{to match} the text description in $\mathcal{P}$. Without the regularizer, the model learns \emph{"coarse matching"} rather than \emph{"fine matching"}. Without fine-matching of $X$ and $\mathcal{P}$, the model lacks the ability to produce exact editing effects following a target prompt $\mathcal{P}^{*}$. Details in \textcolor{blue}{Supplementary}.

\vspace{-1.5mm}
\subsubsection{Temporal Modeling}~\label{sec:temporal}
Enhancing video generation's temporal coherence, the T2I model's self-attention~\cite{rombach2022high} is replaced with cross-frame attention mechanisms~\cite{wu2022tune,Phenaki,ho2022video} (Fig.~\ref{fig:attn_comp}). Options for this include full spatio-temporal attention, causal attention~\cite{Phenaki,ho2022video,he2023latent}, and the computationally efficient sparse-causal attention~\cite{wu2022tune} with $\mathcal{O}((mN)^2)$ complexity, where $m$ typically equals 2.

Simple frame attention, depicted in Fig.~\ref{fig:attn_comp}, suffices for DDIM inversion editing methods since reversed latent features encapsulate temporal information. The implemented attention formula is $\mathrm{Attention}(Q,K,V)=\mathrm{Softmax}(\frac{Q K^T}{\sqrt{d}}) \cdot V$, with 
\begin{equation}
    Q=W^Q z_{i}, K=W^K z_{0}, V=W^V z_{0},
\end{equation}
where $W^Q$, $W^K$, and $W^V$ are trainable projection matrices for query, key, and value, respectively; $z_{i}$ denotes latent features of frame $x_{i}$; and $d$ is the output dimension of key and query features.

Despite sparse-causal attention's superiority in generating videos from noise, its video editing performance falters in rapid-motion scenarios, whereas frame attention benefits from memory efficiency and faster processing (Table~\ref{tab:atten_flops}). Our framework integrates frame-attention with cross-attention for pixel-conditional input correspondence and additional temporal attention for noise prediction in videos~\cite{wu2022tune}. Fine-tuning of the query projection matrices ($W^{Q}$) for both frame and cross-attentions enhances the T2V model's ability to generate semantically consistent, high-quality image sets.

\vspace{-1mm}
\section{Evaluation}
\vspace{-1mm}
\subsection{Implementation Details}
\vspace{-1mm}
Our approach is built upon Latent Diffusion Models~\cite{rombach2022high}, also known as Stable Diffusion, and utilizes the publicly available pre-trained weights. We extract a set of uniformly spaced frames from the input video, each with a resolution of $512\times512$. Subsequently, we fine-tune the model for the 200 epochs method for  iterations, employing a learning rate of $1e^{-3}$ and a batch size of 1. During inference, we utilize the DDIM sampler~\cite{ddim} along with classifier-free guidance~\cite{ho2022classifier} in our experiments. For each individual video, the fine-tuning process requires approximately 3 minutes, while the sampling process takes approximately 20 seconds on an NVIDIA 3090 GPU.

\begin{table}
\centering
\vspace{-1mm}
\scalebox{0.75}{
\begin{tabular}{l|c|c|c}
\toprule
& \multicolumn{2}{c|}{CLIP Score}  &    \\ \cline{2-3}
  \multirow{2}{*}{Method}  & Frame  & Textual   &  \multirow{2}{*}{Avg Edit Time} \\ 
              & Consistency & Alignment & \\ \midrule
CogVideo~\cite{hong2022cogvideo} & 91.25 & 24.40 & N/A \\
Text2LIVE~\cite{text2live}&92.66&26.75& 35 mins. \\  
TAV~\cite{wu2022tune} & 93.89 & 28.86 & 28 mins. \\
Video-P2P~\cite{liu2023video} & 94.58 & 29.15 & 30 mins. \\
\midrule
TAV (LoRA) & 93.52 & 28.46 & 17 mins.\\ 
SAVE w/o Regularizer & 90.35 & 26.15 & 3 mins\\
\textbf{SAVE (Ours)} & \textbf{94.81} & \textbf{29.30} & \textbf{3 mins.} \\ 
\bottomrule
\end{tabular}}
\vspace{-2mm}
\caption{\footnotesize \textbf{Quantitative evaluation} against SOTA baselines.}
\label{tab:quant}
\vspace{-3mm}
\end{table}

\vspace{-1mm}
\subsection{Experimental Results}\label{app}
\vspace{-1mm}\paragraph{Style transfer.} The style of a video is manifested through its comprehensive spatial and temporal attributes. In the second row of Fig.~\ref{fig:main_results}, we introduce the "comic style" to the input video of the rabbit eating the watermelon. It can be observed that our method exhibits the ability to seamlessly transform all frames into the desired style while preserving the semantic content of the original video intact.
\vspace{-1em}
\paragraph{Object Replacement.}
To showcase the efficacy of our method in subject editing, we perform a transformation in the third row of Fig.~\ref{fig:main_results} by replacing  the watermelon with an orange. Similarly, we replace the Jeep with a Porsche car. Notably, our editing outcomes not only align harmoniously with the accompanying text descriptions but also preserve fidelity to the original videos. 
\paragraph{Local Attribute Editing.}By leveraging the generative capabilities of our proposed efficient diffusion generation approach, specific personal attributes can be modified or obfuscated. Fig.~\ref{fig:privacy} showcases personal attributes such as facial features, clothing, or other identifiable characteristics can be edited while maintaining the overall appearance, action information, and realism of the video.


\begin{figure*}[t]
    \centering
  \begin{subfigure}{0.295\linewidth}
    \includegraphics[width=1\linewidth]{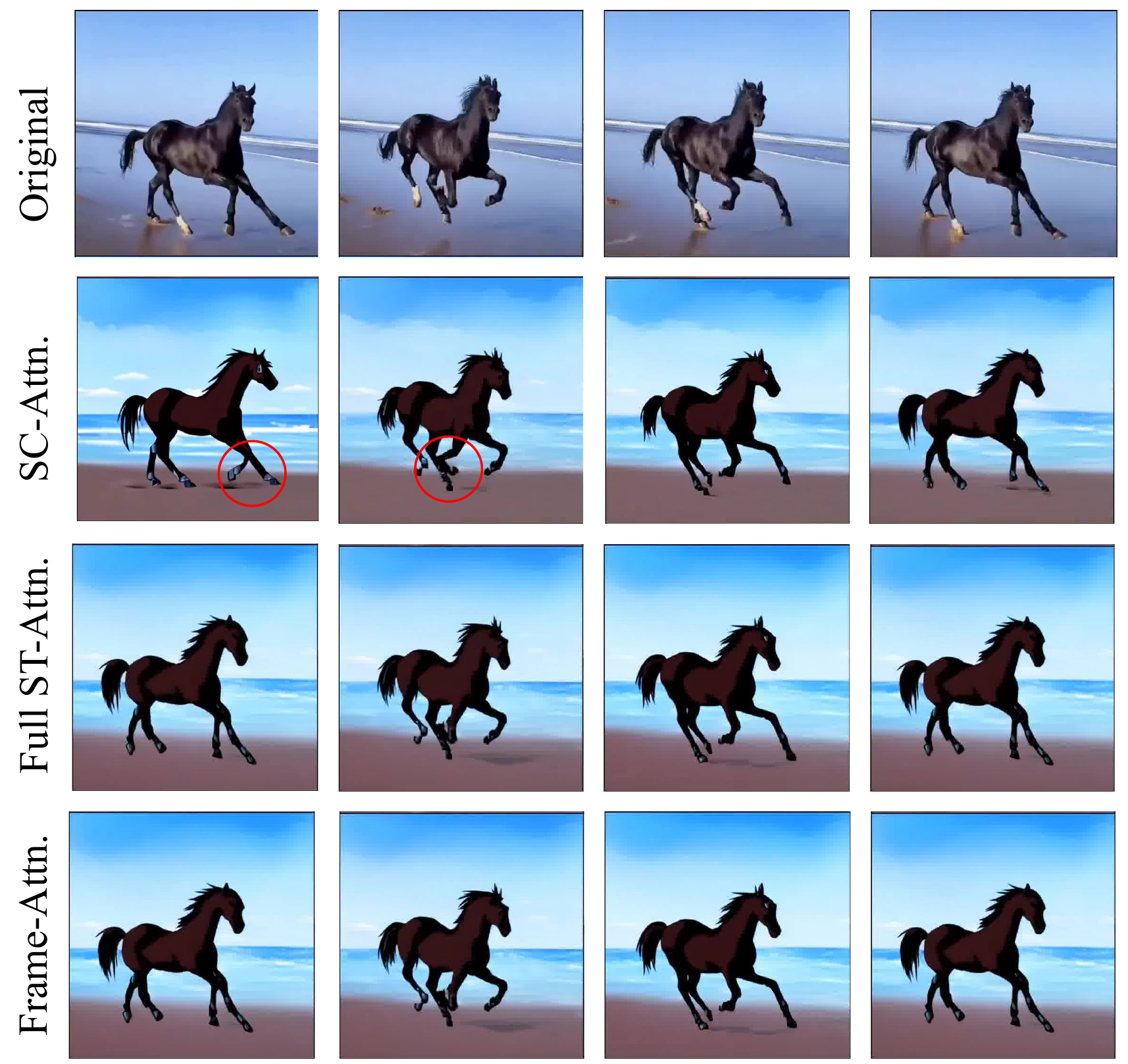}
    \caption{ \scriptsize }
    \label{fig:umar_attn_example}
  \end{subfigure}
  \hfill
    \begin{subfigure}{0.325\linewidth}
    \includegraphics[width=1\linewidth]{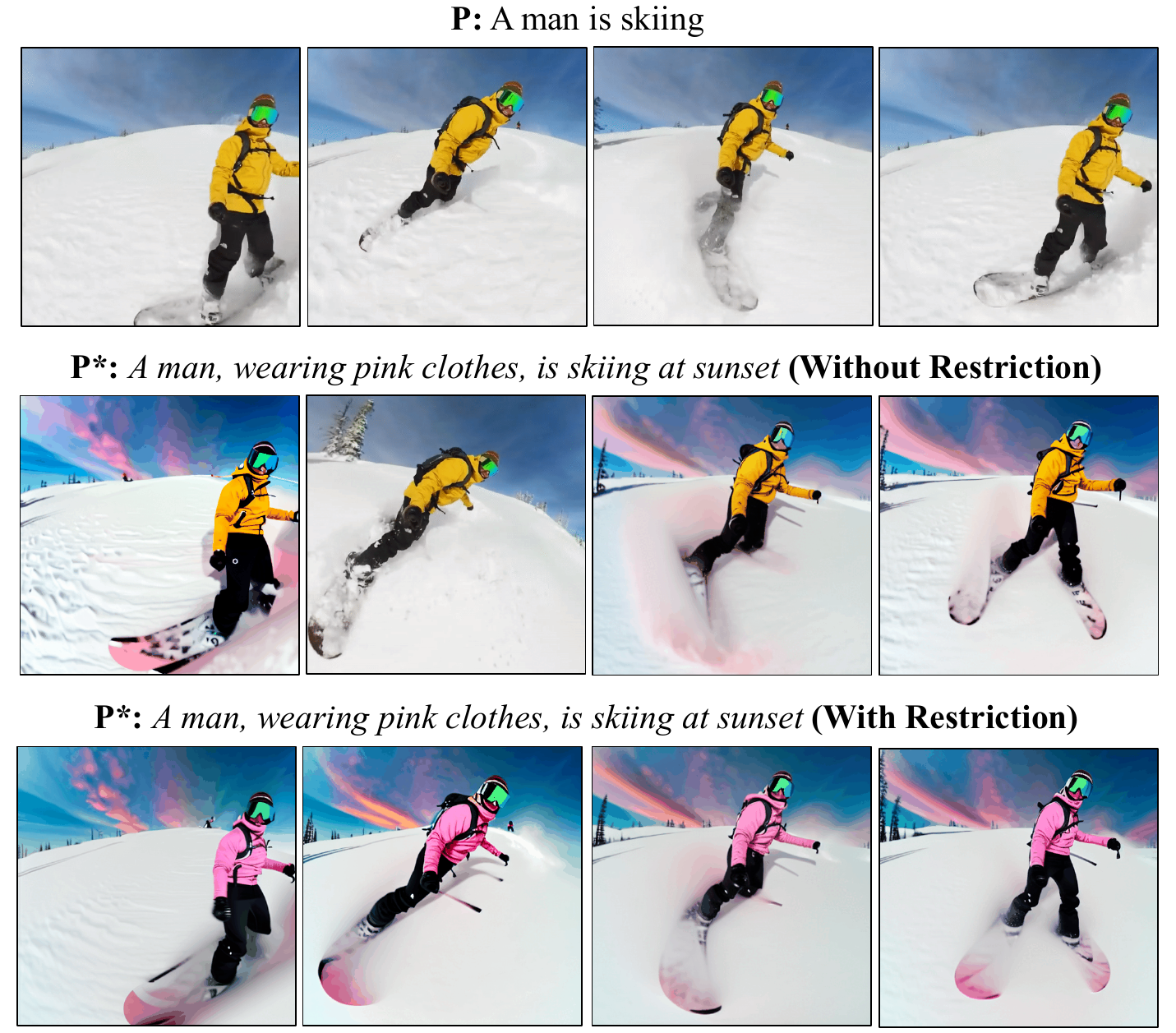}
    \caption{ \scriptsize }
    \label{fig:effect_of_reg}
  \end{subfigure}
  \hfill
  \begin{subfigure}{0.35\linewidth}
    \includegraphics[width=1\linewidth]{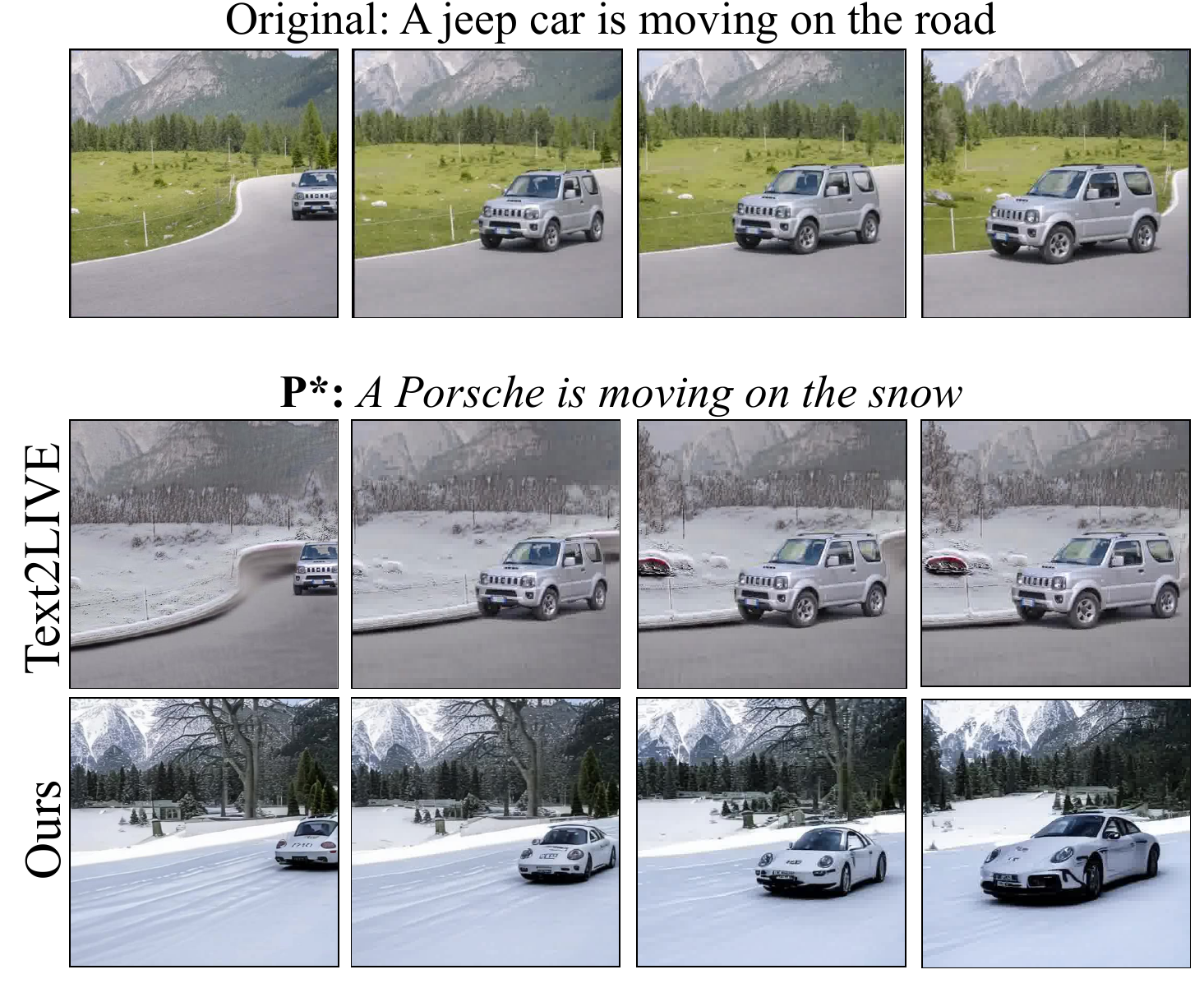}
    \caption{ \scriptsize }
    \label{fig:t2live_comp}
  \end{subfigure}
  \hfill
  \vspace{-2mm}
    \caption{\footnotesize Showcasing the \textbf{a)} results of \textbf{distinct attention mechanisms} against the prompt, "A horse is running on the beach, anime style". \textbf{b)}  fine-tuning performance \textbf{with and without the regularizer or restrictions on the singular values $\Sigma$}. \textbf{c)} superior performance of our method as compared to Text2LIVE which struggles to \textbf{edit shapes}. }
    \label{fig:my_label}
    \vspace{-4mm}
\end{figure*}

\vspace{-1.5em}\paragraph{Multi-attribute Editing.} For multiple attribute editing in the same image, we use a larger regularizer coefficient, $\lambda$ (= 5e-2). We are restricting the large eigenvalues even more for learning finer details which is required here. Figure~\ref{fig:multiple_edits} shows some examples where we are editing multiple attributes successfully. Note that we also need to employ a lower learning rate while training the model for a longer period of time in this case. For $s_{cfg}$, we set a value of 25 here while using a value of 10 for other scenarios. 

\vspace{-1.5em}\paragraph{Runtime Eficciency.} Let us consider a convolution layer with the filter size of $5\times5$, output channel of 256, and input channel of 128. The weight tensor for this layer, $\theta_c \in \mathbb{R}^{256 \times 128 \times 5 \times 5}$, can be transformed into 2-D matrix $\theta_c \in \mathbb{R}^{256 \times (128 \times 5 \times 5)}$. If we take the SVD of this 2D matrix, we only have 256 parameters ($\sigma$) to optimize instead of 8,19,200 parameters. For this particular layer, we reduce the tunable parameter by 3200$\times$ as compared to vanilla fine-tuning. 
By reducing the number of tunable parameters, SAVE significantly improves the computational efficiency. The effect of this can be seen in  Table~\ref{tab:quant} as SAVE significantly reduces the editing time.




\vspace{-1mm}
\subsection{Comparison with Baselines}
\vspace{-1mm}

We compare our method against three baselines: 1) \textit{Tune-A-Video (TAV)}~\cite{wu2022tune} 
2) \textit{CogVideo}~\cite{hong2022cogvideo}, 
3) \textit{Text2LIVE}~\cite{text2live}, 4) Video-P2P~\cite{liu2023video}. Due to space constraints, we have moved certain qualitative comparisons to the \emph{\textcolor{blue}{Supplementary}}.

\vspace{-1em}
\paragraph{Quantitative Comparison.} 
To evaluate the effectiveness of our methodology, we select a total of 42 representative videos sourced from the DAVIS dataset~\cite{pont20172017}. To generate the video content, we employ an off-the-shelf captioning model~\cite{BLIP} as in ~\cite{wu2022tune}. Additionally, we curate a set of 50 edited prompts, specifically tailored to our applications as outlined in Section \ref{app}, through manual design. This meticulous approach ensures comprehensive evaluation across various scenarios and provides valuable insights into the performance and applicability of our proposed approach. Our quantitative evaluation encompasses two main aspects: 1) \emph{CLIP score}, and 2) \emph{Human Feedback}. To evaluate \emph{frame consistency}, we employ CLIP~\cite{radford2021learning} image embeddings to compute the average cosine similarity between all pairs of video frames in the generated output videos. To assess \emph{textual alignment}, we compute the average CLIP score between the frames of the output videos and their corresponding edited prompts (shown in Tab.~\ref{tab:quant}). The detailed study of \emph{Human Feedback} is in the \emph{\textcolor{blue}{Supplementary}}.  

\vspace{-1em}
\paragraph{Qualitative Comparison.}
We present a visual comparison of our proposed approach with baseline methods in Fig.~\ref{fig:comparison}. We observe that Text2LIVE allows for localized area editing, minimizing the overall influence on the rest of the video, but it struggled to edit the black swan into white color. Tune-A-Video lacks the ability to selectively edit specific objects without altering the entire video content as the results against the prompt, "A white swan is swimming on the water" indicate. In contrast, our proposed framework allows for localized area editing, minimizing the overall influence on the rest of the video. Further, our evaluation results reported in Fig.~\ref{fig:t2live_comp} indicate that Text2LIVE struggled to edit the jeep into Porshe. Based on the reported qualitative results, it can be observed that our proposed method generates temporally coherent videos that preserve the structural information from the input video, effectively aligning with the edited words and details. 

\vspace{-1em}
\paragraph{Effect of the Proposed Regularizer.}\label{regularizer}
Note that, our aim is to learn the underlying motion and content of the original video through fine-tuning while editing shape, color, and style using text prompts. This requires better control over the fine-tuning process. Without the regularizer, the model does not learn to fine-match between a text prompt and the video content; which results in limited editability and issues such as unwanted region influence. We show that using cross-attention maps generated with or without the regularizer in Fig.~\ref{fig:effect_of_reg}.  We show a quantitative comparison with and without regularizers in Table~\ref{tab:quant}. It can be observed that the regularizer is very crucial in obtaining the SOTA performance. We provide additional results on this in \emph{\textcolor{blue}{Supplementary}}.

\vspace{-1.5em} \paragraph{Comparison with LoRA.}\label{regularizer} Low-rank adaptation (LoRA)~\cite{hu2021lora} has been initially proposed for efficient fine-tuning of large language models. LoRA operates by adding new tunable rank decomposition matrices into each layer of the network while freezing the original pre-trained model weights. This reduces the number of trainable parameters significantly. Compared to LoRA, our proposed method efficiently modifies the already trained weights without deploying any additional layer of weights. Due to this, SAVE offers better utilization of the full representation power of the original pre-trained weights as compared to LoRA. At the same time, SAVE offers more compact parameter space to optimize. To give an insight into the advantage of compactness, the storage and update requirements for a weight matrix (W) with shape $h \times w$ is O(min(h,w)) for SAVE. On the other hand, it is O(h+w) for LoRA, given a rank of 1. In addition to this, we have a smaller adapted model as compared to LoRA.  Table~\ref{tab:quant} shows the performance of Tune-a-Video with LoRA adaptation layers as compared to SAVE. Note that LoRA may offer better flexibility when one has to perform extensive fine-tuning for learning more than one concept, e.g. few-shot adaptation, or editing multiple videos. By changing the rank, LoRA can adjust its capability according to the task requirement. 

\vspace{-1.5em}
\paragraph{Ablation Study.}\label{ablations}

In order to investigate the impact of temporal modeling design, we conducted additional ablation experiments. Firstly, we examined the removal of dense spatial-temporal attention, which enables bidirectional temporal modeling, and replaced it with Sparse-Causal Attention. The results in the second row of Fig.~\ref{fig:umar_attn_example} demonstrate that the horse running in the initial frames of the video becomes distorted. This distortion primarily arises from the overemphasis on previous frames by the Sparse-Causal Attention, leading to error propagation and significant artifacts in subsequent frames, especially. Across all of our experiments, we observed comparable performance between frame attention with temporal attention in relation to spatio-temporal attention. 

\vspace{-1mm}
\section{Conclusion}
\vspace{-1mm}
We proposed a novel approach for text-guided video editing, adapting an image-based diffusion model by fine-tuning solely the spectral shift within its parameter space. This method significantly reduces the number of adjustable parameters, enhancing computational efficiency. Introducing a spectral shift regularizer, we prioritize constraining larger singular values over smaller ones. This regularization aids the model in capturing nuanced video details aligned with provided text descriptions. Furthermore, we offer theoretical justifications supporting our proposed regularization technique. We support the effectiveness of SAVE through comprehensive evaluations across various scenarios.


{
    \small
    \bibliographystyle{ieeenat_fullname}
    \bibliography{main}
}


\end{document}